\documentclass{svproc}
\usepackage{url}

\input{commands}
\begin{document}
\mainmatter              % start of a contribution
\title{High Dimensional Restrictive Federated Model Selection with multi-objective Bayesian Optimization over shifted distributions}
\titlerunning{Restrictive Federated Model Selection Over Shifted Distributions}  % abbreviated title (for running head)
\author{Xudong Sun\inst{1} \and Andrea Bommert\inst{2} \and Florian Pfisterer\inst{1} \and Jörg Rähenfürher\inst{2} \and Michel Lang\inst{2} \and Bernd Bischl\inst{1}}
\authorrunning{Sun, Bommert, Pfisterer, Rähenfürher, Lang, Bischl} % abbreviated author list (for running head)
\tocauthor{}
\institute{Ludwig Maximilian University of Munich, Munich, Germany \and Technical University of Dortmund, Dortmund, Germany}

\maketitle              % typeset the title of the contribution

\begin{abstract}
A novel machine learning optimization process coined Restrictive Federated Model Selection~(RFMS) is proposed under the scenario, for example, when data from healthcare units can not leave the site it is situated on and it is forbidden to carry out training algorithms on remote data sites due to either technical or privacy and trust concerns. To carry out a clinical research in this scenario, an analyst could train a machine learning model only on local data site, but it is still possible to execute a statistical query at a certain cost in the form of sending a machine learning model to some of the remote data sites and get the performance measures as feedback, maybe due to prediction being usually much cheaper. Compared to federated learning, which is optimizing the model parameters directly by carrying out training across all data sites, RFMS trains model parameters only on one local data site but optimizes hyper parameters across other data sites jointly since hyper-parameters play an important role in machine learning performance. The aim is to get a Pareto optimal model with respective to both local and remote unseen prediction losses,  which could generalize well across data sites. In this work, we specifically consider high dimensional data with different distributions over data sites. As an initial investigation, Bayesian Optimization especially multi-objective Bayesian Optimization is used to guide an adaptive hyper-parameter optimization process to select models under the RFMS scenario. Empirical results shows that solely using the local data site to tune hyper-parameters generalizes poorly across data sites, compared to methods that utilize the local and remote performances. Furthermore, in terms of hypervolumes, multi-objective Bayesian Optimization  algorithms show increased performance across multiple data sites among other candiates.
\keywords{Federated Learning, Multi-objective Bayesian Optimization, High Dimensional Data, Differential Privacy, Distribution Shift, Model Selection}
\end{abstract}

\section{Introduction}
\subsection{Background}
\label{subsec:background}
Federated Learning~\cite{fedlearn, federatelearnCommuni} has drawn increasing attention recently due to overwhelmingly growing data volume and an emerging request for privacy protection from the perspective of individuals, as well as the perspective of data owners, e.g.\ due to GDPR~\cite{melis2018building}.
Usually in federated learning, a server moderates several data sites to carry out optimization iterations, like gradient descent updates, on each data site. Each data site then sends an intermediate result to the server.
The server side aggregates the results and distributes it, so that each data site obtains an updated model. 
This distributed model training process circumvents the bottleneck of data transmission and prevents private data from leaving the data center.
To further increase privacy security against attacks~\cite{melis2018building}, differential private federated learning algorithms have been proposed~\cite{Konecny2016,Truex}.

Current federated learning algorithms rely on an efficient and synchronized communication protocol \cite{federatelearnCommuni, Brendan2017} across the server and different data sites as well as the availability that data on each data site can be used for training. However, it might also be expensive to meet the technical requirements to have a synchronized communication framework needed by federated learning. 

From a privacy protection perspective, several attacks and defenses that undermine privacy in a federated learning context have been proposed \cite{infattacks}, \cite{infattacks2}, \cite{diffpriv}. Differential private federated learning algorithms \cite{Truex,melis2018building, Geyer2018} are based on standard Federated Learning algorithms, with some detail being tailored to fit the need for differential privacy. 

However, there might be restrictions that the data from the remote data site can not be used for training at all. Especially when there is no established trust between parties, privacy protection and attack becomes an arm race, in which case, data owners might want to restrict the access of the data to a maximum extent but still want to participate in the community to build a predictive model that could benefit all sides.  To the best of our knowledge, this is a problem that current differential private federated learning algorithms do not address yet.

In both restricted cases, sending a model to the remote data sites and asking for how good the sent model performs on the remote data sites comes at a certain cost (transmission cost and prediction computation cost for instance). This is comparably acceptable, as only aggregated statistics (typically a single number) need to be reported back.

We coin this new learning scenario Restrictive Federated Learning, emphasizing the point that only data situated locally could be used for model training, while data on the other data sites are partially observed in the sense that the analyst could only observe a scalar performance measure of a sent model on the remote data site, which is restrictive. 

In this restrictive learning scenario, we could only access limited data locally for training a machine learning model, but still want to have a model that could generalize well across the data sites. Therefore, how to do model selection in this special restricted federated learning scenario is of significant interest. 

Bayesian Optimization has proved to be really successful in optimizing machine learning hyper-parameters \cite{snoek2015scalable}. In this work, we want to investigate how it works under the RFMS scenario.

\subsection{Challenges}
A critical challenge in federated learning is unevenly distributed data. For example, there are situations where most features are not available on all data sites~\cite{Konecny2016} or the class distribution is extremely unbalanced across different computation nodes \cite{Zhao}.

In RFMS, there is also the challenge that data can be differently distributed on each data site. 
Specifically, in this work we consider the challenge that distribution of features from one data site might be considerably different from another, due to different sub-populations frequenting a given clinic for example.

Furthermore, the number of observations in clinical research is  usually relatively small, while with the inclusion of genetic data, the number of features can be rather large. This makes model selection~\cite{Guyon2010b,Cawley2010} quite challenging. 
Finding stable predictive models that could generalize well to data collected from different clinical studies or cohorts is difficult.

%In clinical research, usually one prefers models that are parsimonious, sparse, interpretable and with great importance, with a standard software implementation.
%\todo{FP: This is mentioned twice, maybe only mention once and focus}

%To meet the above mentioned challenges, we propose a learning scenario which we call Restrictive Federated Model Selection, 

\section{Problem Statement}
\subsection{Terminology and Notation}
\label{subsec:term}
To clearly address the problem, at the first step, terminologies and notations used throughout the remainder of this paper are explained. 

\textbf{Data site: }Data of a specific domain, clinical research for example, could be located in different places and it is expensive to carry data from one site to another due to technical or privacy concerns. We denote one of such a integrated data unity as a data site.  There is a need to train a specific machine learning model for the domain, which requires collaboration across data sites. We consider data sites of following types.
    
\textbf{Openbox data site $D_{ob}$: }On the openbox data site,  the analyst has full access to the data. A machine learning model can be trained locally using the data situated on openbox data site.  
    
\textbf{Curator data site $D_{cu}$: } From the openbox side, curator data site can be queried for model performance, which can assist the analyst on the openbox data site to get a better model that might generalize across data sites. The curator data site $D_{cu}$ can only be queried with respect to predictive performance, i.e.\ a single aggregate statistic, but the analyst from the openbox side can not access the data in any other way.
This name stems from the field of differential privacy \cite{elder2016bayesian} where there is a curator that controls the data flow which acts like a firewall to $D_{cu}$. 
The curator has full access to $D_{cu}$ but decides on its strategy w.r.t.\ which feedback value to give to the statistical query by actively perturbing and coordinating the answers given to the queries. In this work, we assume a honest answer to the query except otherwise specified.

\textbf{Lockbox data site $D_{lb}$: } Lockbox \cite{Gossmann2018} data site refers to data sites which the analyst from the openbox side can not access by any means. In practice, lockbox correspond to data sites that could not contribute in the process of building a machine learning model due to various reasons, but are likely to participate in the future or simply benefit from the model built. From a model evaluation perspective, $D_{lb}$ on the other hand could measure how good a machine learning model generalizes to completely unseen data.
    
\textbf{Inbag and outbag: } For evaluation purposes, we hold out a fraction (say 20\%) of the curator and openbox data which we call \textbf{outbag}, denoted by $D_{cu}^{og}$ and $D_{ob}^{og}$, the leftover is called \textbf{inbag}, which is $D_{cu}^{ig}$ and $D_{ob}^{ig}$. For simplicity, we use $D_{ob}$ to represent $D_{ob}^{ig}$ when the context is about learning and use $D_{ob}$ to represent $D_{ob}^{og}$ when the context is about evaluating how good a method is. Also, we define the inbag and outbag of lockbox to be identical to lockbox itself, i.e. $D_{lb}=D_{lb}^{ig}=D_{lb}^{og}$.
%\paragraph{\textbf{Lockbox inbag and outbag:} }  

 \textbf{Model parameter $\theta$ and hyper-parameter $\phi$:} A machine learning algorithm, given a dataset $D_l$, where $l$ means "learn" or "local", $D_l = D_{ob}^{ig}$, for example, and a set of hyperparameters $\phi$, learns a model specified by a set of model parameters $\theta = \hypDtom(D_{l}\mid \phi)$ where $\hypDtom$ represent the learning process to map a dataset $D_l$ associated with a set of hyper-parameter $\phi$ to a set of machine learning model parameter $\theta$. 
 
\textbf{Model performance and loss:} 
The performance of a model characterized by $\theta$ to a data site~$D$ is given by $$\perf \left( D \mid \theta = \hypDtom(D_l\mid \phi)\right)$$ where $\perf$ computes an estimate of predictive performance on $D$, under model parameter $\theta$ trained from dataset $D_l$, based on hyper-parameter $\phi$. By convention, we use  $J$ to represent a regret that need to be minimized, which could be $1-accuracy$ for example.

%\paragraph{\textbf{Levels of Parameter}} To make an analogy with Bayesian machine learning algorithms, we define $\theta$ to be the first level parameter, and $\phi$ the second level parameter.

\textbf{Restricted Federated Model Selection (RFMS) Scenario: } The analyst from the openbox side want to initiate a study to a specific domain(clinical studies like cancer research for example). A machine learning model that fits the data well on the openbox side, as well as one that could generalize to a certain extent across the other data sites is required. Due to privacy sensitivity or technical difficulty, some data sites could only collaborate in a model selection process in the form of curators. Each query to the curator from the openbox side is at a certain cost. Note that all forms of data sites including openbox, curator and lockbox should be used to evaluate the selected model whenever possible. 
%We use $\theta$ to denote the model parameter, for example, the linear coefficients in a support vector machine model or the split points in a random forest model. 
%Then the modeling fitting process could be written as a Maximum-a-Posterior estimation $\theta^{*}(D_{ob},  \phi) = argmax_{\theta} p(\theta|\phi; D_{ob})$, where we use the $\phi$ to denote the hyper-parameters, for example, the $C$ and $Sigma$ parameters of the RBF kernel support vector machine.

\subsection{An example of RFMS on high dimesional unevenly distributed data}
Gene Expression Omnibous (GEO) is a public available functional genomics data repository with array and sequence based data that researchers from around the world could contribute to. Although the data in GEO is publicly available instead of privacy sensitive, the origin that the datasets in GEO comes from different sources makes it a perfect example of RFMS.
We use the breast cancer datasets GSE16446, GSE20194, GSE20271, GSE32646, and GSE6861 from the GEO database \cite{edgar2002gene, mccall2010frozen}. Each dataset we consider here could be regarded as a data site due to the fact that they come from different sources, by different contributors.

The publicly available microarray gene expression datasets were accessed via tools provided by the Gene Expression Omnibus (GEO) data repository. %(\url{http://www.ncbi.nlm.nih.gov/geo/}; accessed June 2015). 
Frozen robust multiarray analysis (fRMA) \cite{mccall2010frozen} was used for normalization.
All breast cancer datasets were checked for duplicates and a pair of patients was considered duplicate when the correlation of their expression values was at least 0.999. 
Duplicates were removed. 
The response variable is binary (classes "pathological complete response" and "residual disease") for all datasets.
The six observations with a missing value for the response variable are omitted.
The resulting numbers of observations per dataset are displayed in Table~\ref{tab:geo_data_distr}.
\begin{table}[ht]
    \centering
    \begin{tabular}{l|rrrrr}
         GEO-ID & 16446 & 20194 & 20271 & 32646 & 6861 \\
         Observations &      114 &     211  &   178    &  115  & 161 \\
    \end{tabular}
    \caption{Number of observations per GEO dataset}
    \label{tab:geo_data_distr}
\end{table}
The datasets contain clinical and gene expression data.
We do not consider the clinical variables because many values are missing.
The gene expression data has been measured on three different types of microarray chips (HG-U133-Plus2 for GSE16446 and GSE32646, HG-U133-A for GSE20194 and GSE20271, and HG-U133-X3P for GSE6861).
As the measured genes differ between the three chips, we only consider the genes that are measured on all of the chips.
Out of these 1965~genes, we only use the 1000~genes with the highest variances across all patients and datasets.

\begin{figure}[ht]
		\centering
		\includegraphics[scale=0.30]{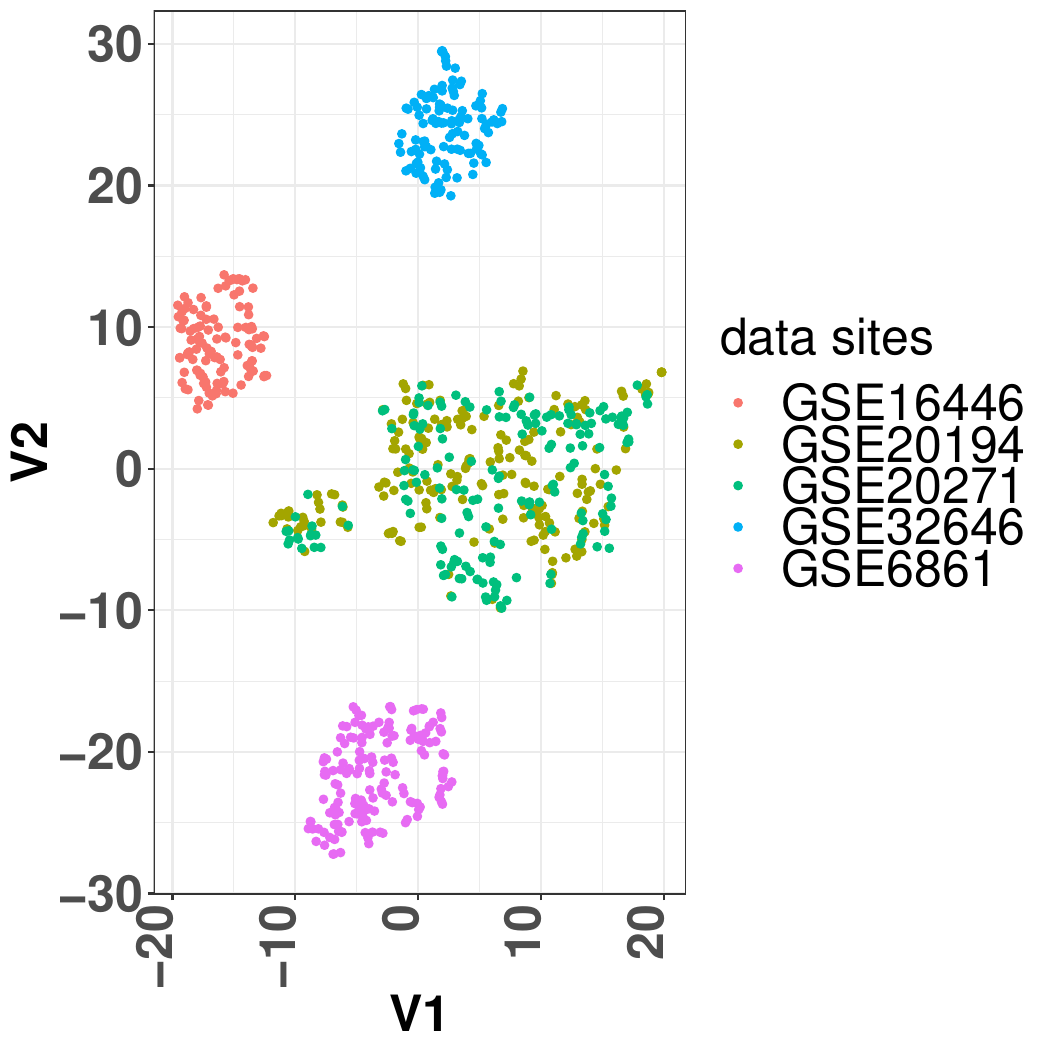}
		\caption{T-SNE plot for the GEO datasets over data sites.}
		\label{fig:geo-tsne}
\end{figure}

It can be assumed that the relation between the response variable and the covariates is not identical across the datasets and the features distribution also varies from data site to data site. This is typical for gene expression data, especially if it has been measured on different chips, at different times, at different places and after different times until the tissue was frozen.
A T-SNE~\cite{VanDerMaaten2008} plot by pooling the feature part of the data together from these data sites can be found in Figure~\ref{fig:geo-tsne} where the colors indicate different data sites. From Figure~\ref{fig:geo-tsne}, it is obvious that the data sites lie on different locations in the low dimension embedding, which is a clear indicator of distribution shift across data sites. We will use this example as a major case in this paper. 
\unskip\vspace*{-7pt} 
\unskip\vspace*{-7pt}
\subsection{Evaluation Criteria}
\label{subsec:eval}
To further explain the problem, 
before discussing any potential solution, we first address the question of how to evaluate model performance, which will help deeper understanding of the problem.

%In RFMS, it is not simply a training set, validation set and test set, which is the classical concept when data are all located in one place or at the full grasp of the analyst. 
%Conventional model selection process of machine learning algorithm using model based optimization on the hyper-parameter space splits one dataset into train validation and test part(Nested Cross Validation) for example, which tunes performance only on the validation dataset. 
In RFMS, we want to obtain a model that genaralize well for the openbox, curator and hopefully for the lockbox as well, which is a multi-objective problem. Accordingly, the selected model should also be evaluated with method that could take different objectives into consideration.

%Proposal:
%In our scenario, we can not resort to the traditional split int train, validation and test set. 

%We argue that this is a multi-objective problem 
%since an optimal model should work good for the openbox, curator and lockbox simultaneously.

\textbf{Dominated hypervolume: } A natural criterion is to measure the Dominated Hypervolume\cite{beume2006faster} of the model performance on the outbag part of openbox and curator site, as well as the lockbox, as in Equation (\ref{eq:hv})
\begin{align}
&J^{hv}(\phi \mid D_{ob}^{og},~D_{cu}^{og},~D_{lb})=\hyperv\left[f_{ob}^{og},~f_{cu}^{og},~f_{lb} ~\right],\nonumber \\
%\text{where,} \nonumber \\
&f_{ob}^{og} = \perf \left( D_{ob}^{og} \mid \theta =\hypDtom(D_{ob}^{ig}\mid \phi)\right), \nonumber \\
&f_{cu}^{og} = \perf \left( D_{cu}^{og} \mid \theta =\hypDtom(D_{ob}^{ig}\mid \phi)\right), \nonumber \\
&f_{lb} = \perf \left( D_{lb} \mid \theta =\hypDtom(D_{ob}^{ig}\mid \phi)\right).
\label{eq:hv}
\end{align}
where, $\hyperv$ represent the calculation of the Dominated Hypervolume, and the performance on each data site outbag part is represented as $f_{ob}^{og},~f_{cu}^{og},~f_{lb}$ respectively. Dominated Hypervolume Indicator is also known as Lebesgue Measure or S-Metric which is the hypervolume between a non-dominated front and a reference point. Due to space limit, we invite readers who are not familiar with these multi-objective concepts to refer to the references. 

\section{Related Work}
In this section, we review recent works that has connections with RFMS.

\textbf{Nested Cross Validation(NCV):} NCV \cite{Guyon2010b} uses an outer loop cross validation to safe guard the risk of overfitting during the hyper-parameter tuning process. However, RFMS does not allow cross validation due to the  constraint that remote data site can not be used for training.

\textbf{Federated Learning:}
Federated learning \cite{fedlearn} also consider situations where data is distributed non-i.i.d. across several data sites and possibly unbalanced, but they assume scenarios where data is fully accessible over a huge amounts of data sites compared to a smaller number of data points available at each site. This is different from RFMS, where we consider data can only be accessed through prediction. Moreover, in RFMS, we consider a relatively small amount of data sites with less instances but high dimensional data.

\textbf{Distribution Shift: } 
Distribution Shift refers to a mismatch in distribution between the data an algorithm was trained on, and data used for model validation or prediction.
Detecting and characterizing such shift remains an open problem~\cite{ZhangSMW2013,Rabanser2018}. In this work, we do not drive deeper in theory of the data shift problem, but provides an empirical study which partially addresses the data shift problem, especially when feature distribution varies across data sites.

\textbf{Train On Validation: }
In \cite{Tennenholtz} the authors use parts of the validation dataset for training to generate a stable algorithm. In \cite{Zeng2017}, a progressive resampling process is used. However, both works assume that all the data in question is available for training, which is not possible in RFMS.

\textbf{Thresholdout Family: } \cite{Dwork2017} showes that differential privacy is deeply associated with model generalization and propose the Thresholdout algorithm to avoid overfitting on the validation set due to repetitive usage. \cite{Gossmann2018} extends the instance wise Thresholdout to AUC measures. However, these methods rely on the i.i.d assumption of data which does not fit our scenario here.

\textbf{Adaptive Regularization: }
In \cite{Rendle2012}, the author proposed an alternative update method for model parameter $\theta$ and hyper-parameter $\phi = \lambda$ of a recommendation system\cite{kushwaha2018lesson}, where the $\lambda$ is the regularization parameter. In adaptive regularization, the update for the $\lambda$ is based on the "future" value of performance which is also similar to the EM algorithm update process.  However, adaptive regularization only works with gradient based algorithms. Especially, it is only implemented for Factorization Machine in libFM. So in general it does not work for non-gradient based optimization typed machine learning models. 

\textbf{Model Agnostic Meta Learning (MAML): }
Model Agnostic Meta Learning~\cite{Finn2017a} originates from few shot learning. It aims at adapting to new instances, in which sense is similar to RFMS. However, MAML works only with gradient based method and pre-assumes that the algorithm could see the full subsequent dataset which is not possible in RFMS problem setting.
\section{Methods} 
In this section, we first describe the general RFMS process in \ref{subsec:rfms}, then in \ref{subsec:boms}, we propose how to handle the RFMS process with Bayesian Optimization. 

\pgfdeclarelayer{-1}
\pgfsetlayers{-1,main}
\tikzset{zlevel/.style={%
    execute at begin scope={\pgfonlayer{#1}},
    execute at end scope={\endpgfonlayer}
}}

\tikzset{every picture/.style=thick}
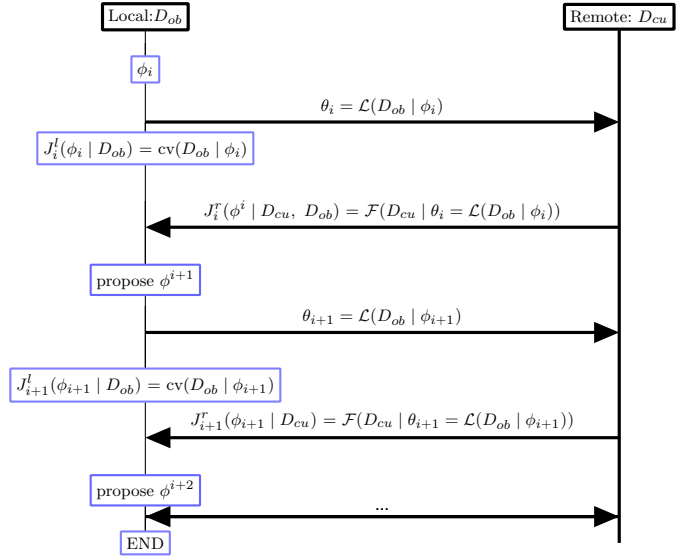
\begin{figure}[h]
	\centering
\begin{tikzpicture}[
scale=0.7, every node/.style={scale=0.7}
,every node/.append style={very thick,rounded corners=0.1mm}
]
%\begin{pgfonlayer}{tp}    % select the background layer
\node[draw,rectangle] (local) at (0,0) {Local:$D_{ob}$};
\node[draw,rectangle] (Server) at (9,0) {Remote: $D_{cu}$};
\begin{scope}[zlevel=-1]
\draw [very thick] (Server)--++(0,-10);
\end{scope}
\begin{scope}[zlevel=main]
\node[draw=blue!50,rectangle,thick] (phi0) at (0,-1) {$\phi_{i}$};
\node[draw=blue!50,rectangle,thick] (cv) at (0,-2.5) {$J_i^{l}(\phi_{i} \mid D_{ob})$ = cv($D_{ob} \mid \phi_{i}$)};
%\end{pgfonlayer}
\node[draw=blue!50,rectangle,thick] (cv2) at (0,-7) {$J_{i+1}^{l}(\phi_{i+1}\mid D_{ob})$ = cv($D_{ob}\mid \phi_{i+1})$};
\node[draw=blue!50,rectangle,thick] (end) at (0,-10) {END};
\node[draw=blue!60, rectangle,thick] (propose) at (0,-5) {propose $\phi^{i+1}$}; %J_i^{l}(D_{ob},~\phi_{i}),J^r_i(D_{cu},~\phi^{i}) 
\node[draw=blue!60, rectangle,thick] (propose2) at (0,-9) {propose $\phi^{i+2}$};
\end{scope}
\begin{scope}[zlevel=-1]
\draw [] (local)  --  (phi0)--(0,-0.5);
\draw [] (phi0)  --  (cv)--(0,-2); % -2 is absolute vertical position of the starting point of this line
\draw [] (cv)  --  (propose)--(0,-3.5);
\draw [] (propose)  --  (cv2)--(0,-5.5);
\draw [] (cv2)  --  (propose2)--(0,-9.5);
\draw [] (propose2)  --  (end)--(0,-9.5);
\end{scope}
\draw [->,very thick] (0,-2)--node [auto] {$\theta_{i} =  \hypDtom(D_{ob}\mid\phi_{i}) $}++(9,0);

\draw [<-,very thick] (0,-4)--node
[auto] {$J^r_i(\phi^{i}\mid D_{cu},~D_{ob})$ = $\perf(D_{cu}\mid \theta_{i} = \hypDtom(D_{ob}\mid \phi_{i}))$}++(9,0);

\draw [->,very thick] (0,-6)--node [auto] {$\theta_{i+1} =  \hypDtom(D_{ob}\mid \phi_{i+1}) $}++(9,0);

\draw [<-,very thick] (0,-8)--node [auto] {$J^r_{i+1}(\phi_{i+1}\mid D_{cu})$ = $\perf(D_{cu}\mid \theta_{i+1} = \hypDtom(D_{ob}\mid \phi_{i+1}))$}++(9,0);

\draw [->,very thick] (0,-9.5)--node [auto] {...}++(9,0);
\draw [<-,very thick] (0,-9.5)--node [auto] {...}++(9,0);
\end{tikzpicture}
\caption{Restrictive Federated Model Selection starting from step $i$}
\label{fig:pr-method}
\end{figure}

\subsection{Restrictive Federated Model Selection}
\label{subsec:rfms}
The general process of RFMS is illustrated in Figure \ref{fig:pr-method}, which depicts an asynchronous communication process during optimization. At step $i$, based on hyper-parameter $\phi_{i}$, the machine learning model is trained on $D_{ob}$ to get the model parameter $\theta^{i} =  \hypDtom(D_{ob}\mid \phi^{i})$. 

With the same hyper-parameter $\phi_{i}$, a $10$-fold cross validation is carried out on the openbox inbag part $D_{ob}$, which gives us one loss function in Equation (\ref{eq:jl}).

\begin{equation}
J_i^{l}(\phi_{i}\mid D_{ob}^{ig}) = cv(D_{ob}^{ig}\mid \phi_{i})
\label{eq:jl}
\end{equation} 
where $cv(D_{ob}^{ig}\mid \phi_{i})$ represent the average loss of the cross validation and $J^l_i$ means local loss at the $i$th step.

Another loss function is obtained by sending the model parameters $\theta^{i}$ to the remote side as shown in Equation (\ref{eq:jr})
\begin{equation}
J^r_i(\phi_{i}\mid D_{cu}^{ig})=\perf(D_{cu}^{ig}\mid~\theta_{i}=\hypDtom(D_{ob}^{ig}\mid \phi_{i}))\label{eq:jr}\end{equation}
Here $J^r_i$ means loss on the remote curator at the $i$th step. 

At the next step, a decision process $\beta$ (see Algorithm \ref{algo:fmobo-ms}) based on all historical observations will propose a new hyper-parameter to be tried out for a potential better performance. This process is repeated until budget reached. The process should return the optimal hyper-parameters. The complete procedure is listed in Algorithm \ref{algo:fmobo-ms}, where the the decision process $\beta$ to generate the proposal is approximately greedily taking the optimal of a Gaussian Process originated surrogate $\mu(\phi \mid \rep,~\Phi)$, Expected Improvement \cite{snoek2012}, for instance. We use $\Phi$ (with an inital design sized $n^{ini}$) to represent the hyper-parameter buffer and $\rep$ to represent the corresponding objective(s) buffer.
%Secondly, our model selection dataset $D_{cu}$ lies in some remote sources which we can not easily grasp so that we can not use them for fitting the model.

\subsection{Bayesian Optimization and Baselines}
\label{subsec:boms}

Bayesian optimization tries to solve the problem of optimizing (often expensive-to-evaluate) black-box functions by using an internal empirical performance model which learns a surrogate model of the objective function while optimizing it. 
A widely used application for Bayesian Optimization \cite{jones1998} is the optimization of hyperparameters \cite{snoek2012,Bergstra2011} of machine learning algorithms.
Its aim is to find an optimal configuration $\phi^\star$ from the feasible region. %$\Phi$, where each $\phi \in \Phi$ usually consists of multiple parameters that need to be optimized jointly. 
The choice of hyperparameters for a machine learning model influences the learned model and can thus result in different performances (cf. \cite{Rijn2018,Probst2018}).
%cf., an abbreviation for the Latin word confer (the imperative singular form of "conferre"), literally meaning "bring together", is used to refer to other material or ideas which may provide similar or different information or arguments.

Since the distribution of the data across different data sites is unknown, we propose to treat the model selection approach as a black box optimization problem. 
Specifically, we use Bayesian Optimization in Algorithm \ref{algo:fmobo-ms} to solve the Restrictive Federated Model Selection problem with the following variants.
%\paragraph{Cheating Single Objective(cso)} To assess to what extent tuning has worked, we pool the openbox and curator data sites to form our local dataset $D_{local} = (D_{ob}, D_{cu})$ which we denote cso (cheating single objective).

\textbf{Local Single Objective (lso) Bayesian Optimization:} In local single objective (lso) Bayesian Optimization, we
set objective function as cross validation performance on the local openbox data site, hyper-parameters are tuned based on $J^{lso}(\phi) = J^{l}= cv\leftb D_{ob}^{ig}\mid\phi\rightb$
where $J^l$ is defined in Equation (\ref{eq:jl}).

\textbf{Federated Single Objective (fso) Bayesian Optimization: }
In Federated Single Objective Bayesian Optimization, we combine the openbox cross validation aggregated results in Equation (\ref{eq:jl}) and curator performance in Equation (\ref{eq:jr}) linearly as objective function, hyper-parameters are tuned based on 
\begin{align}
 &J^{fso}(\phi) = \alpha~J^{l}(\phi_{i}\mid D_{ob}^{ig})+ \left( 1-\alpha \right)J^r(\phi_{i}\mid D_{cu}^{ig})  \nonumber \\ 
 &\alpha \in \left[0,~1\right].
\end{align}%\quad 
Specifically, we use $fso2$ to represent $\alpha = 0.2$ and $fso8$ to represent $\alpha =0.8 $ and so on. Note that $\alpha = 1$ corresponds to $lso$. We use different $\alpha$ to check if there is an obvious effects by changing $\alpha$.

\textbf{Federated Multiobjective Objective ($\mathbf{fmo}$) Bayesian Optimization : }
Multiobjective Bayesian Optimization \cite{horn2017first} optimizes multiple objectives simultaneously, by random linear combination or optimization a S-metric based objective, which avoid deciding which linear combination parameter $\alpha$ to choose. In this work, we use the Parego algorithm \cite{Knowles2006ParEGOAH}  to optimize the local objective in Equation (\ref{eq:jl}) and remote objective in Equation (\ref{eq:jr}) jointly.

\textbf{Random Search Multiobjective (\randmo): } To evaluate whether Bayesian optimization makes sense, we randomly search the hyper-parameter space and select the pareto front \cite{van1998evolutionary}  as final output, 
which we call random search multi-objective.

\begin{algorithm}[t]
\caption{RFMS with Bayesian Optimization (RFMS-BO)}\label{algo:fmobo-ms}
\begin{algorithmic}[1]
\Procedure{RFMS-BO}{} \Comment{data site notation here refer to the \textbf{inbag} part}
\State $\Phi_{1:n^{ini}} = \left\{\phi_1, \hdots, \phi_{n^{ini}}\right\}$  \Comment{initial design as hyper-parameter buffer}
\State $\rep_0= \emptyset$ \Comment{objective  buffer}
\For{$i$ in $1:n^{ini}$, $\phi_i$ in $\Phi_{1:n^{ini}}$} 
\State $J_i^{l}(\phi_{i}\mid D_{ob}) = cv(D_{ob}\mid \phi_{i})$  \Comment{Cross validation performance aggregation as loss}
\State $\theta_i = \hypDtom(D_{ob}\mid\phi_i)$\Comment{training on $D_{ob}$ with $\phi_i$}
\State $J^r_i(\phi_{i}\mid D_{cu}, D_{ob})$ = $\perf(D_{cu}\mid \theta_{i})$ \label{algoline:perf} \Comment{test on curator}
\State $\rep_{i} = \rep_{i-1} \concate \lefts J_i^l, J_i^r \rights$ \Comment{populate objective buffer}
\EndFor
\State fit $\mu(\phi \mid \rep_i,~\Phi_{1:n^{ini}})$  \Comment{train Surrogate Function}
\State $j = i + 1$
\While{budget not reached}
\State $\phi_{j}= \beta(\mu(\phi \mid \rep_{j-1},~\Phi_{1:j-1}))$ \Comment{propose new hyper-parameter}
\State ${\Phi}_{1:j} = \Phi_{1:j-1}  \concate \lefts \phi_{j} \rights$  \Comment{populate hyper-parameter buffer}
\State $J_j^{l}(\phi_{j}\mid D_{ob}) = cv(D_{ob}\mid \phi_{j})$
\State $\theta_j = \hypDtom(D_{ob}\mid \phi_{j})$
\State $J^r_j(\phi_{j} \mid D_{cu})$ = $\perf(D_{cu}\mid \theta_{j})$  \label{algoline:perf2}
\State ${\rep}_{j} = \rep_{j-1}  \concate \lefts J_j^l,~J_j^r \rights$  \Comment{populate objective buffer}
\State $j \gets j+1$
\State update $\mu(\phi \mid \rep_j,~\Phi_{1:j})$ \Comment{update surrogate}
\EndWhile\label{euclidendwhile}
\State $i^{*} = \argmax_{i}(\rep)$
\State $\{\phi^{*}\} =  \Phi_{i*}$
\State $\{\theta^{*}\} = \hypDtom (D_{ob};\phi^{*})$ 
\State \textbf{return} $\phi^{*}, \theta^{*}$ 
\EndProcedure
\end{algorithmic}
\end{algorithm}

%We bootstrap the out-bag part of open-box and curator with alpha preference to open-box, we use this bootstrapped version as the open-box vs curator evaluation.
%
%We then bootstrap the open-box out-bag part and entire lock-box with alpha preference, thus we get the open-box vs lock-box performance. 

%\paragraph{Ensemble}
%Ensemble of the Pareto models might generate a stronger model which might not be so fair to our baseline model. So it is not considered. 
\subsection{Semi-simulation of Data sites} 
\label{sec:semi-sim}
Publicly available datasets which could fit into the RFMS scenario intrinsically are rare. To get data from a diversified source aside from the Gene Expression Ominbus, we turn to approximate the RFMS scenario by splitting an existing dataset into different parts as if each part sits on a different data site. In practice, we always split an existing dataset into $5$ parts to keep consistence with our GEO datasets. 

Since we use real data, but kind of simulate to split the dataset into different data sites to fit into the RFMS scenario, we call this semi-simulation of data sites. We propose the following strategy to semi-simulate the data sites. 

\textbf{Stratified Random Split (SRS): }  First, split the dataset into two parts according to a factor column. Specifically, we use the target column in a classification dataset. Then, each factor part is randomly split into $5$ buckets. The positive class part got $b^p_1, \hdots, b^p_5$ and the negative class part got $b^n_1, \hdots, b^n_5$, where $b^n_i$ and $b^p_i$ represent the $i$th bucket in the negative part and positive part respectively. Lastly, sort the buckets in each factor part according to the number of instances and combine the buckets in reversing order to form each data site, i.e., $d_i = b^n_{s^n(i)} \concate b^p_{s^p(6-i)}$, where $d_i$ represents the $i$th combined data site, $s^n$ and $s^p$ are the sorted index vector of each part. We use $\concate$ to denote pooling two data buckets.

\textbf{Dimension Reduction and Clustering(DRC): }  
First, carry out a dimension reduction technique on the dataset like Principal Component Analysis. Then split the dataset into positive class part and negative class part. Cluster each part into $5$ clusters, .i.e. $c^n_1,~\hdots,~c^n_5$ for the negative class part and $c^p_1,~\hdots,~c^p_5$ for the positive class part. Sort the clusters with respect to the cluster size in each part and combine them in reversed order to form each data site, i.e., $d_i=c^n_{s^n(i)} \concate c^p_{s^p(6-i)}$, where $d_i$ represent the $i$th combined data site, $s^n$ and $s^p$ are the sorted index vector of each part. We use $\concate$ to denote pooling two data together.

We choose Mixture of Gaussian Model (MOG) for the clustering, due to consideration that MOG could also serve as a density estimator. 

\begin{equation}
p(X)=\mathlarger{\mathlarger{\Sigma}}_{k=1}^5 c_k \mathcal{N}(X| \boldsymbol{\mu}_k, \boldsymbol{\Sigma}_k)
\label{eq:mog}
\end{equation}

In MOG, each cluster is represented by a Gaussian distribution $\mathcal{N}(X| \boldsymbol{\mu}_k, \boldsymbol{\Sigma}_k)$ with its own parameters $\boldsymbol{\mu}_k$(mean) and $\boldsymbol{\Sigma}_k$(covariance), as shown in Equation (\ref{eq:mog}), $c_k$ is the mixing coefficient of each cluster. For each of the chosen datasets in Table \ref{tb:ds}, we model the data distribution as $p(X)$ in Equation(\ref{eq:mog}) and approximately, each cluster resulted data site represent a different distribution. For simplicity, we assume all clusters are with different mean vectors but share the same covariance matrix to assemble a distribution shift. The T-SNE plot is done to the SRS scenario (Figure \ref{fig:tsne-stratif}) and the DRC scenario. In DRC, we use PCA as dimension reduction, keeping 10\% (Figure \ref{fig:tsne-pca1}) and 50\% (Figure \ref{fig:tsne-pca5}) of the total variance to tell if the reduced dimension makes a big difference in generating an unevenly distributed data sites scenario). From these figures, we do not observe a big difference between different percentage of variance to reserve in PCA, but observe a big difference between SRS and DRC where SRS generates a more evenly distributed data sites, while DRC generates more uneven distributions across different clusters(data sites).
%\mfig{fig/graph/pca.png}{variance distribution of oml 3891}{pca}{0.5}
\begin{figure}[ht]
	\begin{minipage}[b]{.3\textwidth}
		\mgraph{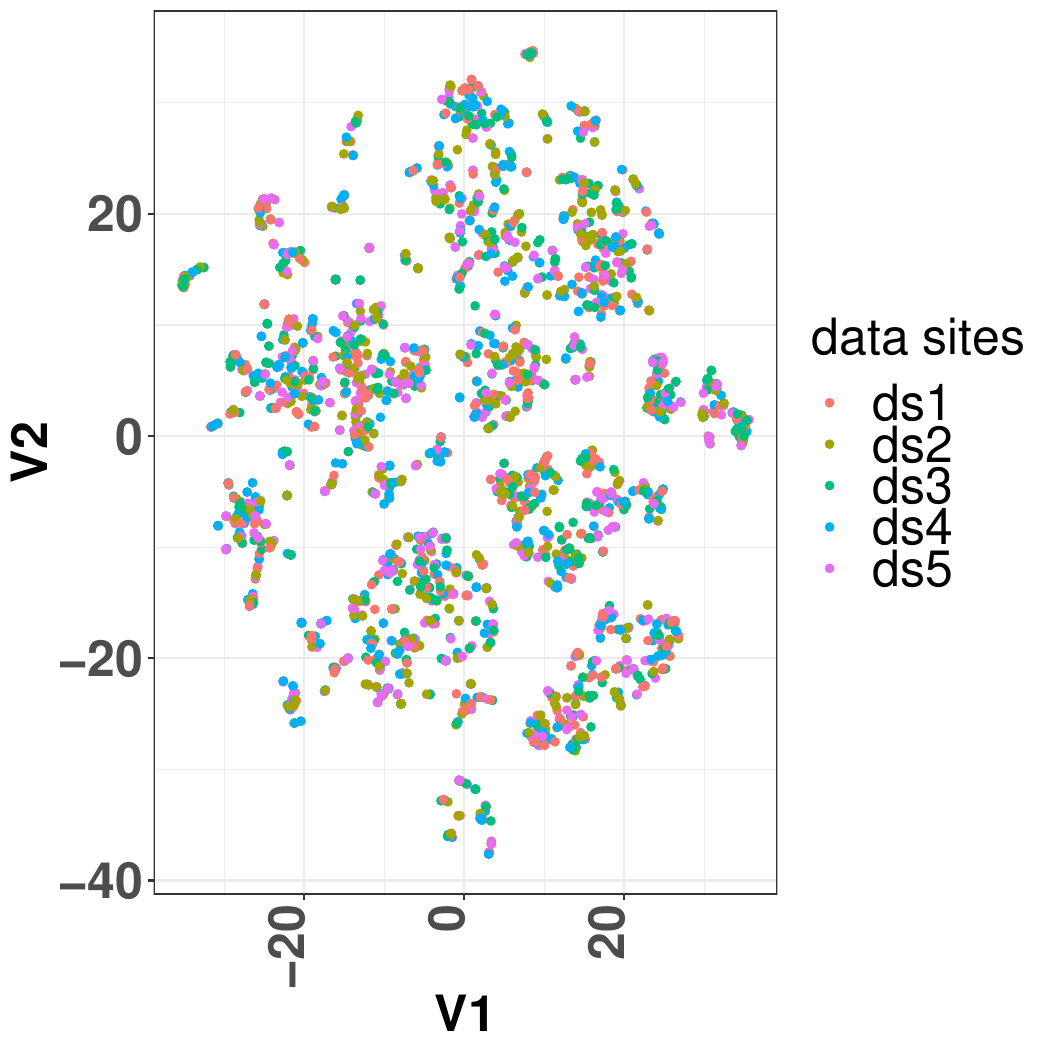}{Stratified Random Split (SRS)}{fig:tsne-stratif}{0.3}
	\end{minipage}
	\hfill
	\begin{minipage}[b]{.3\textwidth}
		\mgraph{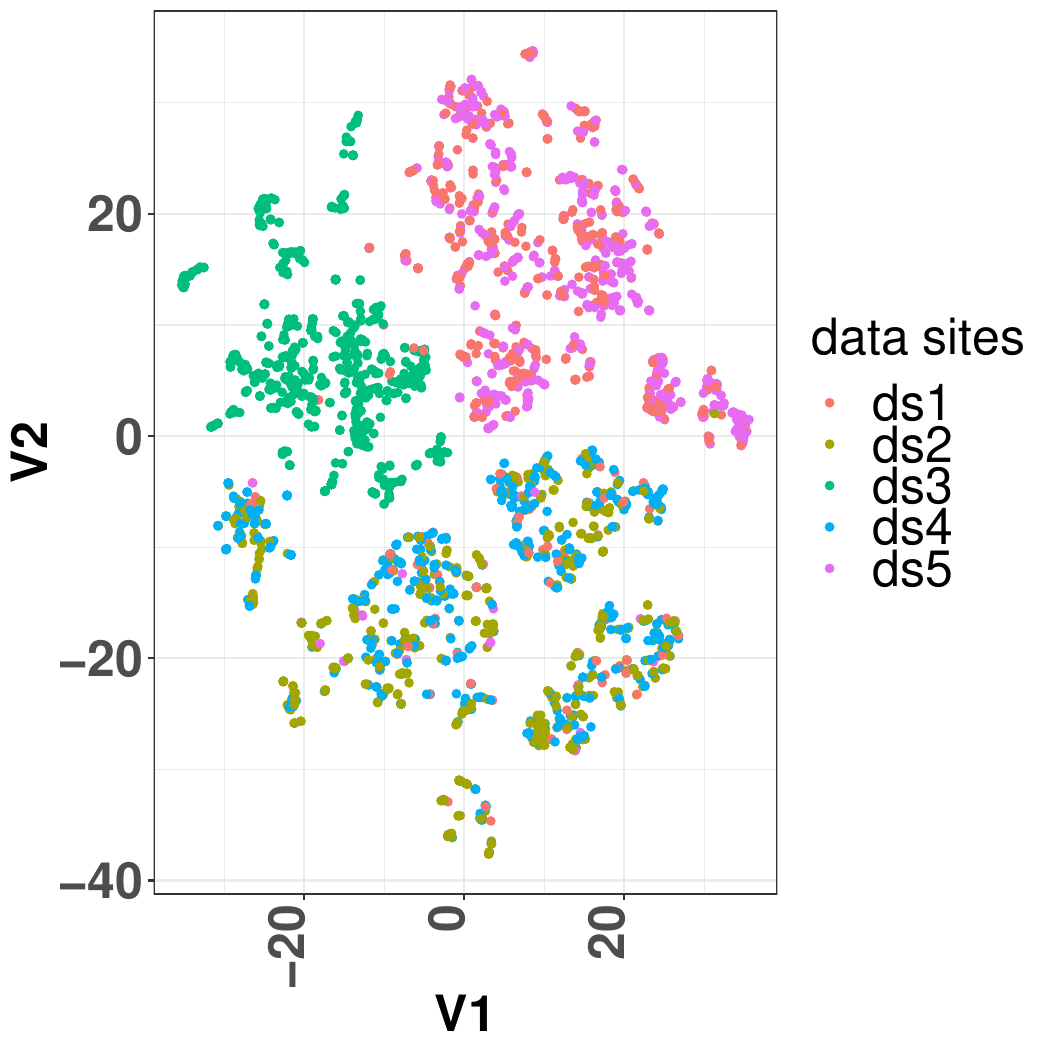}{DRC with PCA and keep 10\% variance}{fig:tsne-pca1}{0.3}
	\end{minipage}
	\hfill
	\begin{minipage}[b]{.3\textwidth}
		\mgraph{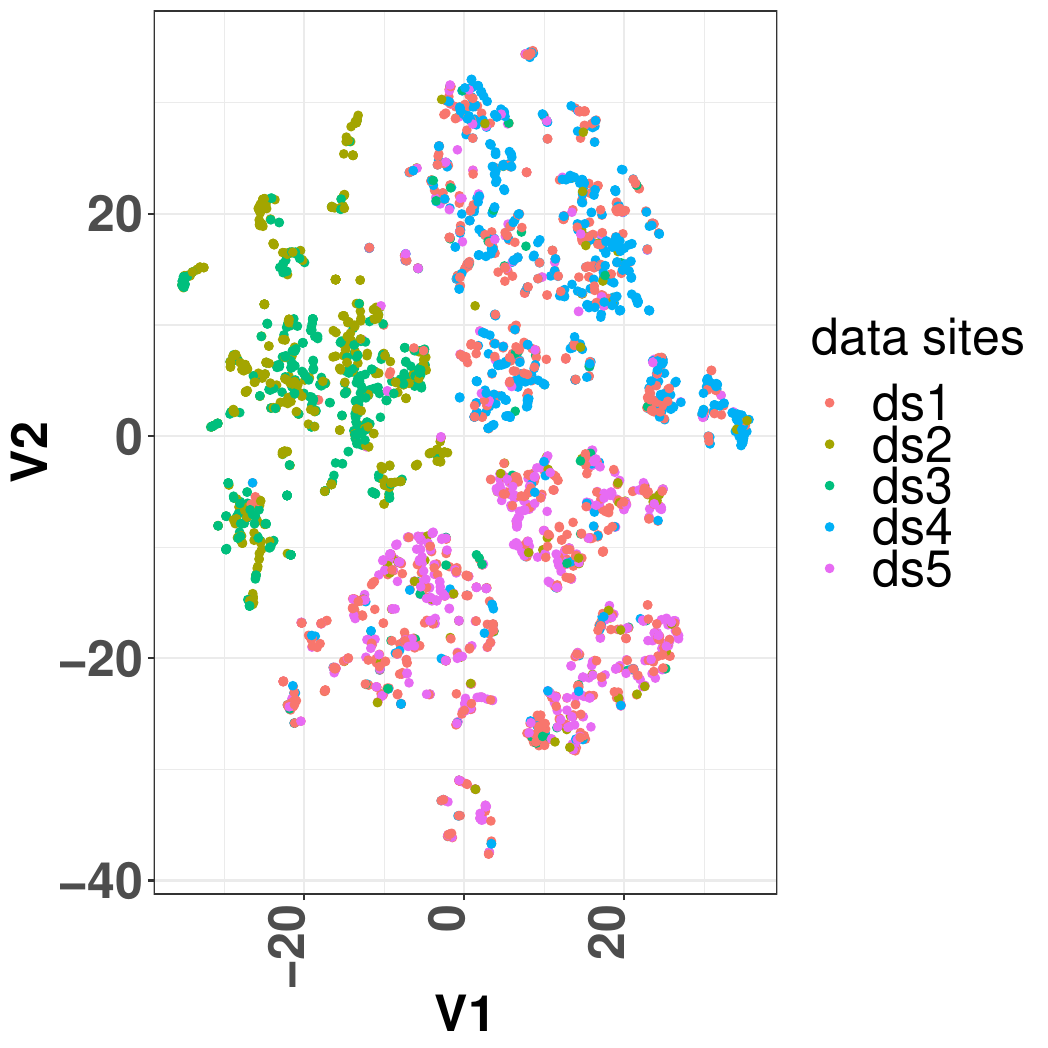}{DRC with PCA and keep 50\% variance}{fig:tsne-pca5}{0.3}
	\end{minipage}
\end{figure}

\section{Experiment}
\subsection{Settings}
\label{exp:setting}
Since we have selected $5$ datasets from the Gene Expression Ominibus to represent $5$ data sites, we  will consider the exemplary problem of $5$ data sites for the remainder of the paper.

%As shown in Figure \ref{fig:ds-div}, 
In the experiment, one of the $5$ data sites is used as openbox $D_{ob}$, another one as lockbox $D_{lb}$ and the three left over are used as curators $D_{cu}$. 
We choose to have only one openbox to simulate the scenario, that usually only local data at the current data sites are fully available to the analyst. We choose to have $3$ curators and only $1$ lockbox to simulate the scenario that more data sites want to collaborate with the openbox data site. Curator data site losses are weighted by the size of the each curator data site during optimization.
With this strategy, there are in total $ 5 \times 4 = 20$ combinations of openbox-curator-lockbox on the $5$ datasets. 
%Figure \ref{fig:ds-div} shows $4$ out of the $20$ combinations when lockbox is fixed to be the $5th$ data site.  
Each openbox and lockbox combination defines one scenario. Each scenario is repeated $10$ times ($10$ replications) where we call each replication one experiment. 
% mitigate the effect of which data site get randomly assigned to the openbox or lockbox.
We sequentially run all RFMS methods, described in Section~\ref{subsec:boms}, with 3 machine learning algorithms (kernel support vector machine, random forest and elastic net). Thus, we have in total $20 \times 10 \times 3 = 600$ experiments given a RFMS problem with $5$ data sites. All Bayesian Optimization procedures share the same initial design of 20 randomly selected configurations, and are then run for another $40$ iterations.  Thus in total we have a budget of $60$ evaluations. To have a fair comparison, Random Search use the same number of evaluations.

In order to evaluate our method, we randomly partition openbox and the curator into two parts, namely an inbag part(80\%) and an outbag part(20\%). Replications mentioned above could average out the random splits and other stochastic factors. We use $D_{ob}^{ig}$ for training a model, and use $D_{cu}^{ig}$ as well as $D_{ob}^{ig}$ for model selection. The outbag parts of openbox and curator are reserved for post-hoc analysis. This allows us to assess, whether our methods overfit in each of the two boxes. Additionally, performance is also recorded on the lockbox site for another aspect of evaluation. We then compare the different methods described in Section~\ref{subsec:boms} on the outbag portion of the respective boxes (as noted in \ref{subsec:term}, all data of lockbox belongs to outbag).\urldef\urlsourcecode\url{https://github.com/compstat-lmu/paper_2019_multiobjective_rfms}\footnotemark\footnotetext{source code in \urlsourcecode}

\subsection{Selection of Dataset for semi-simulation}
In order to validate our results on different data sources, we obtain additional data sets from OpenML \cite{Vanschoren2014}. As no datasets with an intrinsic splitting mechanism such as the GEO dataset (where each dataset comes from a particular source) are available, we simulate the RFMS scenario according to the strategies described in Section~\ref{sec:semi-sim}.

Model generalization becomes more difficult when there are comparatively more features than instances. Therefore, we restrict ourselves to datasets with a relatively high-dimension characteristics: Since we intend to split a dataset into $5$ parts as $5$ data sites, the number or instances in each data site is approximately reduced by $5$ times compared to the original dataset (we rebalanced cluster results which generate too small clusters but adding instances to the smallest cluster from the biggest cluster until the smallest cluster reaches 10 percent of the total number of instances), but the number of features over the number of instances get to be approximately $5$ times of the original ratio, so a $p$ (number of features) over $n$ (number of instances) ratio of more than $0.2$ in the original dataset corresponds to $\frac{p}{n}=1$ in each data site, thus we consider datasets with $\frac{p}{n}$ ratio around $0.2$ to be high-dimensional.

Too few instances is more prone to problems in data resampling processes like cross validation. For example, one fold of the cross validation might contain no instance from the underrepresented class. Thus we do not want too extremely unbalanced classification datasets. In order to have a sufficient amount of data in each of the 5 boxes, we select only data sets with more than $500$ instances. For the purpose of simplicity, we additionally restrict our data set selection to data sets that are $i)$ binary class, $ii)$ do not have missing values. As a result, we use the data sets in Table \ref{tb:ds} to provide additional validation of the proposed methods.
 \begin{table}[ht]
 \centering
 \caption{List of datasets from OpenML}
%\begin{small}
%\begin{sc}
%\resizebox{\columnwidth}{!}{%
 \begin{tabular}{rllrrr} %%lcccr
   \hline
   &name & n & p & p/n & class ratio \\ 
   \hline
   &gina agnostic & 3468 & 970 & 0.28 &  0.97 \\ 
   &Bioresponse & 3751 & 1776 & 0.47  & 0.84 \\ 
   &$fri\_c4\_500\_100$\footnotemark & 500 & 100 &0.2 &0.77\\
    \hline
 \end{tabular}
%}
%\end{sc}
%\end{small}
\label{tb:ds}
 \end{table}
\footnotetext{https://www.openml.org/d/742}  
\newcommand\polygon[5][]%
{   \pgfmathsetmacro{\angle}{360/#2}
    \pgfmathsetmacro{\startangle}{-90 + \angle/2}
    \pgfmathsetmacro{\y}{cos(\angle/2)}
    \begin{scope}[#1]
        \foreach \i in {#4}
        {   \pgfmathsetmacro{\x}{\startangle + \angle*\i}
          \fill[#5] (0, 0) -- (\x:#3) --  (\x + \angle:#3) -- cycle;
        }
        \foreach \i in {1,2,...,#2}
        {
          \pgfmathsetmacro{\x}{\startangle + \angle*\i}
          \draw (0, 0) -- (\x:#3) --  (\x + \angle:#3) -- cycle;
        }
    \end{scope}
}

%\begin{figure}
%    \centering
%    \begin{tikzpicture}[scale=0.6, every node/.style={scale=0.6}
%,every node/.append style={very thick,rounded corners=0.1mm}]
%    \polygon{6}{1}{1,4}{cyan}
%    \polygon[xshift=3cm]{7}{1}{1,2,4,7}{cyan}
%    \polygon[xshift=8cm]{8}{3}{5,6,1,4}{cyan}
%\end{tikzpicture}
%\caption{class distribution pertaining dataset division}
 %   \label{fig:my_label}
%\end{figure}

\subsection{Machine learning algorithms and hyper-parameters}
We choose $3$ machine learning algorithms(which we call learner) based on the consideration that the learners should be representative to different mechanisms of various machine learning algorithms. Elastic net logistic regression (implemented in R package $glmnet$ \cite{friedman2009glmnet}) is a good representative for linear classifier which could deal with high dimensional data(\textbf{classif.glmnet}), thus chosen because according to \cite{Strang}, one should not rule out simple models prematurely.
R package $ranger$ \cite{wright2015ranger} implements a random forest(\textbf{classif.ranger}) which is a state of art non-linear learner that has shown outstanding performance. 
Kernel support vector machine ($ksvm$)(\textbf{classif.ksvm}) implemented in \cite{karatzoglou2004kernlab}, is a nonlinear classifier which could deal with high dimensional data. 
\urldef\urlmlrglmnet\url{https://github.com/mlr-org/mlr/blob/3edac9f65ed5c157a3d868fe8d2908eaa2a09ebd/R/RLearner_classif_glmnet.R#L7}
The hyper-parameters to be optimized with their ranges are shown in Table \ref{tb:hyperpar}. Hyper-parameter tuning is done with $mlr$\cite{bischl2016mlr} and $mlrMBO$\cite{bischl2017mlrmbo}. Meaning of hyper-parameters can be found in respective packages. \footnotemark \footnotetext{\urlmlrglmnet}
%\todo{AB: Add references for the learners.}
%\todo{AB: From the names of the hyperpars it is not easy to tell what the meaning of the hyperpar is.}
%Algorithms like $glmnet$ does not seem as a strong but there are of great importance in the clinical field due to interpretability and a standar implementation.

%We are aware of the recent advances in lasso-based feature selection methods \TODO{Add some citation here} but it will be future work to evaluate their performance once there is better implementation. 
%\vspace*{-10pt}
\begin{table}[h]
	\centering
	\caption{List of Hyperparameters}
	%	\footnotesize
	%	\resizebox{\columnwidth}{!}{%
	\begin{tabular}{lllr}
		\toprule
		\textbf{Classifier}&\textbf{Hyperparameter}&\textbf{Type}&\textbf{Range}\\
		\midrule
		glmnet&                  alpha&	    numeric&  ($0$,$1$)\\
		glmnet&                  s&	    numeric&  ($2^{-10}$,$2^{10}$)\\
		ksvm&                  C&	    numeric&  ($2^{-15}$,$2^{15}$)\\
		ksvm&              sigma&	    numeric&  ($2^{-15}$,$2^{15}$)\\
		random forest&          num.trees&	    integer&            (100,5000)\\
		random forest&      min.node.size&	    integer&                (1,50)\\
		random forest&    sample.fraction&	    numeric&               (0.1,1)\\
		% classif.xgboost&                eta&	    numeric&           (0.001,0.3)\\
		% classif.xgboost&         max\_depth&	    integer&                (1,15)\\
		% classif.xgboost& min\_child\_weight&	    numeric&                (0,50)\\
		\bottomrule
	\end{tabular}
	%	}
	\label{tb:hyperpar}
\end{table}
\subsection{Results and Discussion}
%\mfig{fig/graph/violin_geo_count.pdf}{inbag outbag geo dataset}{box}{0.50}
%\mfig{fig/graph/openbox-curator-alpha.pdf}{openbox--curator-alpha-plot}{box}{0.50}
In this section, we compare different candidates of RMFS methods proposed in section \ref{subsec:eval} with respect to their predictive performance. 
Our aim is to obtain machine learning models, that generalize well across data sites. As an aggregate measure, we choose the dominated hypervolume of the data kept out-of-bag in the openbox $D_{ob}^{og}$, curator $D_{cu}^{og}$ and lockbox $D_{lb}$ respectively as shown in Equation (\ref{eq:hv}). We consider the average performance on the curators for calculating the hypervolume. Lockbox data measures how our methods generalize to sub-populations not considered at all during the training and model selection process. 
Using hypervolume results in a comprehensive overview of them.
%\vspace*{-7pt}
\subsubsection{Results on the GEO datasets}
As shown in Figure \ref{fig:hv}, we compare the mean dominated hypervolume from Equation (\ref{eq:hv}) of 3 machine learning algorithms (corresponding to the 3 panels in the plot) and several RFMS methods. We aggregate over $10$ replications and $20$ combinations of possible openbox-lockbox combinations. 

From Figure~\ref{fig:hv}, we can observe that $lso$ performs the worst among other candidates, showing that in the RFMS scenario, solely tuning hyper-parameters on the local openbox data site will usually not lead to a model that generalizes well across data sites, which is in accordance with intuition.  
The other candidates methods including \fmo\, and several \fso\, variants, that predicting on the data of the curator and using this performance as a feedback performs better, showing that the feedback could help in arriving at models which generalize better. However, the considered Bayesian Optimization approaches do not overrate the multi-objective random search \randmo, nor do we observe any effect of changing $\alpha$ in the performance of \fso\,.
%\mfig{fig/graph/geo_box.pdf}{geo}{geo}{0.5}
%\mfig{fig/graph/mean_hypervolumes_geo.pdf}{Dominated hypervolume on GEO datasets}{hv}{0.3}{h}
%\mfig{fig/graph/wins_and_losses_geo.pdf}{Comparison of wins and losses on GEO dataset}{win-geo}{0.35}{h}
\begin{figure}[ht]
	\begin{minipage}[b]{.45\textwidth}
		\mgraph{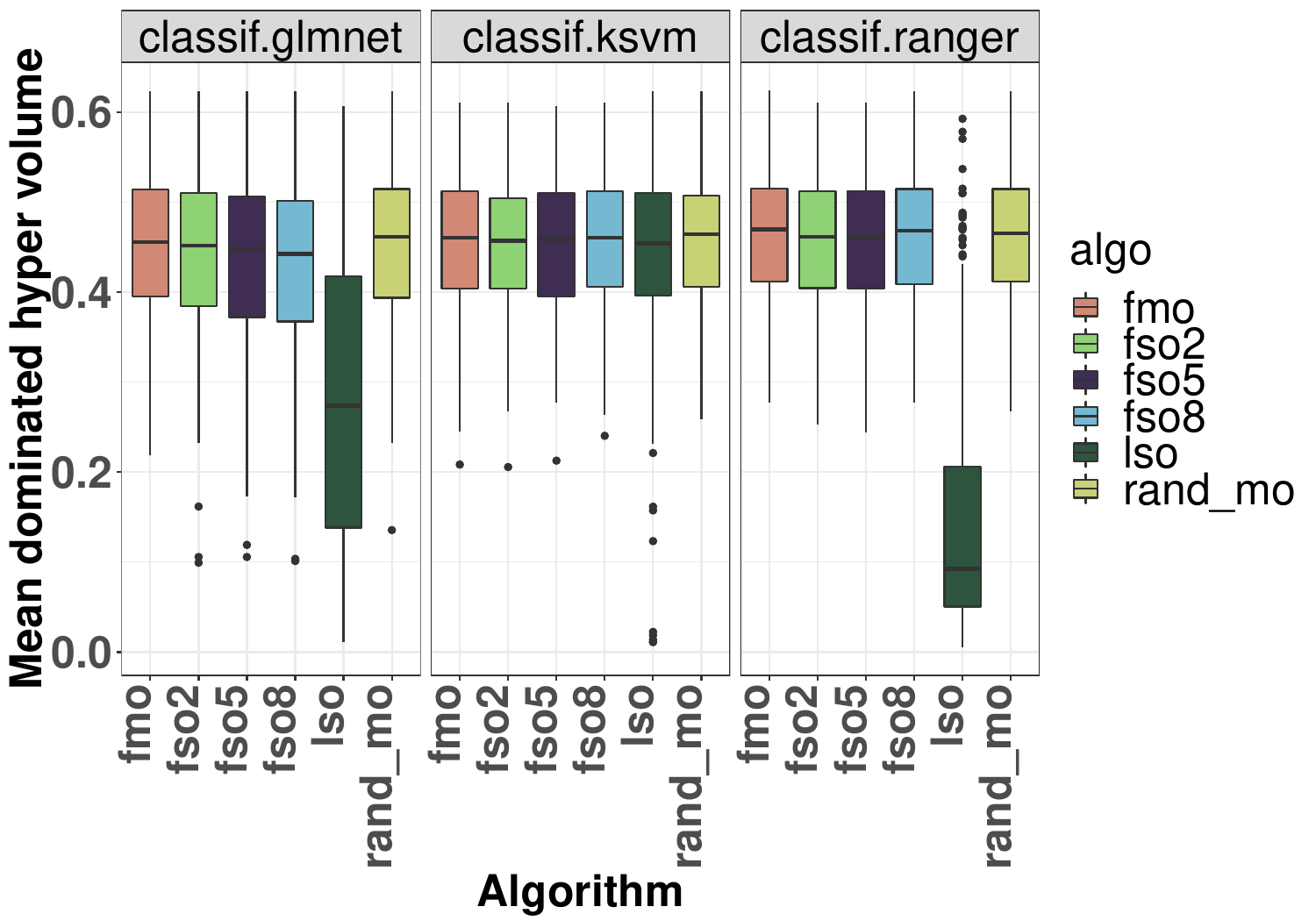}{Dominated hypervolume on GEO datasets}{fig:hv}{0.3}
	\end{minipage}
	\hfill
	\begin{minipage}[b]{.4\textwidth}
		\mgraph{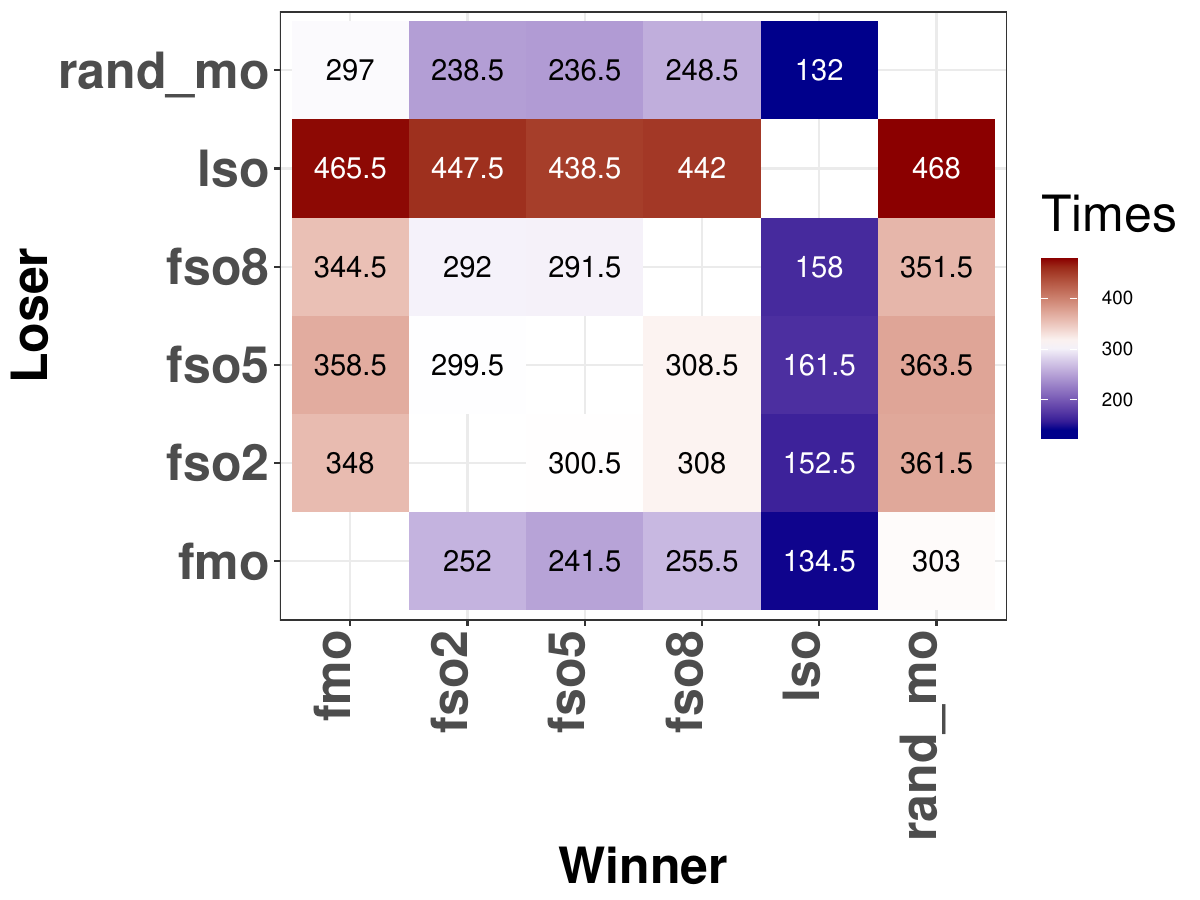}{Comparison of wins and losses on GEO dataset}{fig:win-geo}{0.3}
	\end{minipage}
\end{figure}
In order to make a more precise comparison, we compare the pairwise wins and losses of all the RFMS methods in terms of dominated hypervolume. For each experiment, we build a $0-1$ matrix to compare the win and loss of each algorithm pair (when method A is compared against method B, we take 0 for loss, 1 for win, and $0.5$ for tie)  and aggregate the matrix across all $600$ experiments. Results are shown in Figure \ref{fig:win-geo}, where the horizontal axis corresponds to winners and the vertical axis correspond to losers. The elements in the matrix correspond to how many times the winner has won against the loser. It is easily observable that both bi-criteria methods (\fmo\, and \randmo) are slightly better than other candidates, as they win more than half of the experiments.
%\vspace*{-45pt}

%\vspace*{-35pt}
%\setlength{\belowcaptionskip}{-10pt}
\subsubsection{Results on the semi-simulated RFMS scenario}
To avoid single dataset bias, we also analyze how the same algorithms compare under our semi-simulated RFMS scenario described in section \ref{sec:semi-sim} over data of various sources. 

\textbf{Dimension Reduction and Clustering (DRC): }
We first simulate the RFMS scenario with DRC explained in section \ref{sec:semi-sim}, which could result in a situation that data from different data sites are differently distributed, where we keep 10 percent variance in the PCA step.

Figure \ref{fig:hv-oml-comb-pca} shows the dominated hypervolume by aggregating across all the datasets in Table \ref{tb:ds}. Compared to Figure \ref{fig:hv}, it is more obvious here that the multi-objective methods work better than the single objective Bayesian optimization methods. In Figure \ref{fig:oml-comb-pca-win}, we have the Winner-vs-Loser plot for the aggregated results on the OpenML datasets listed in Table \ref{tb:ds}, where the multi-objective candidates outperform the rest by a large margin. Furthermore, \fmo\, wins \randmo\, by a considerable margin, giving confidence that Bayesian Optimization make a difference compared to random search.

%\mfig{fig/graph/mean_hypervolumes_oml_combined_pca_3sets.pdf}{Aggregated mean dominated hypervolume under DRC scenarios obtained over OpenML datasets}{hv-oml-comb-pca}{0.3}{}

%\mfig{fig/graph/wins_and_losses_oml_combined_pca_3sets.pdf}{Aggregated wins and losses on the DRC scenarios obtained over OpenML datasets}{oml-comb-pca-win}{0.35}{}

\begin{figure}[ht]
	\begin{minipage}[b]{.5\textwidth}
		\mgraph{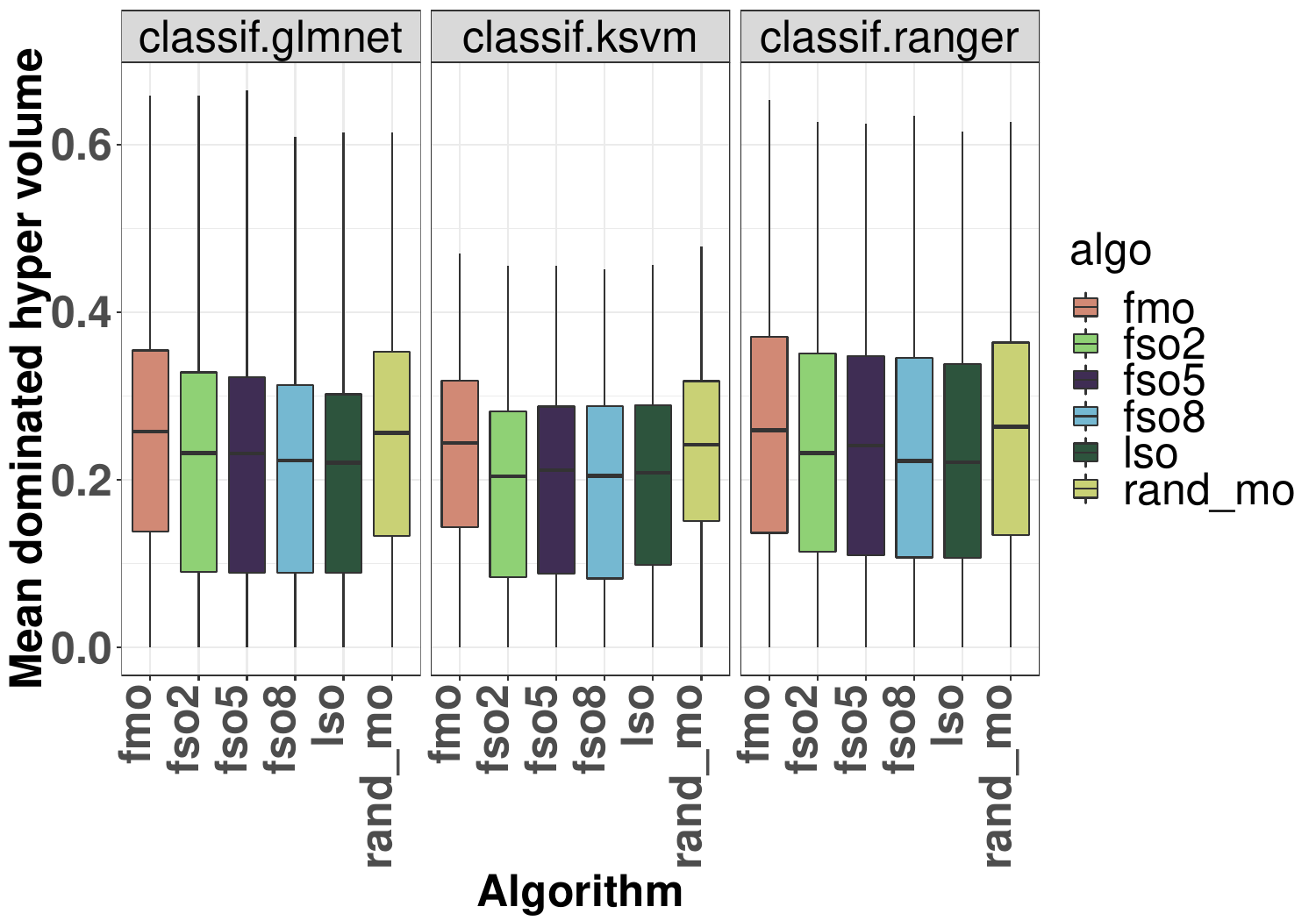}{Aggregated mean dominated hypervolume under DRC scenarios obtained over OpenML datasets}{fig:hv-oml-comb-pca}{0.3}
	\end{minipage}
	\hfill
	\begin{minipage}[b]{.4\textwidth}
		\mgraph{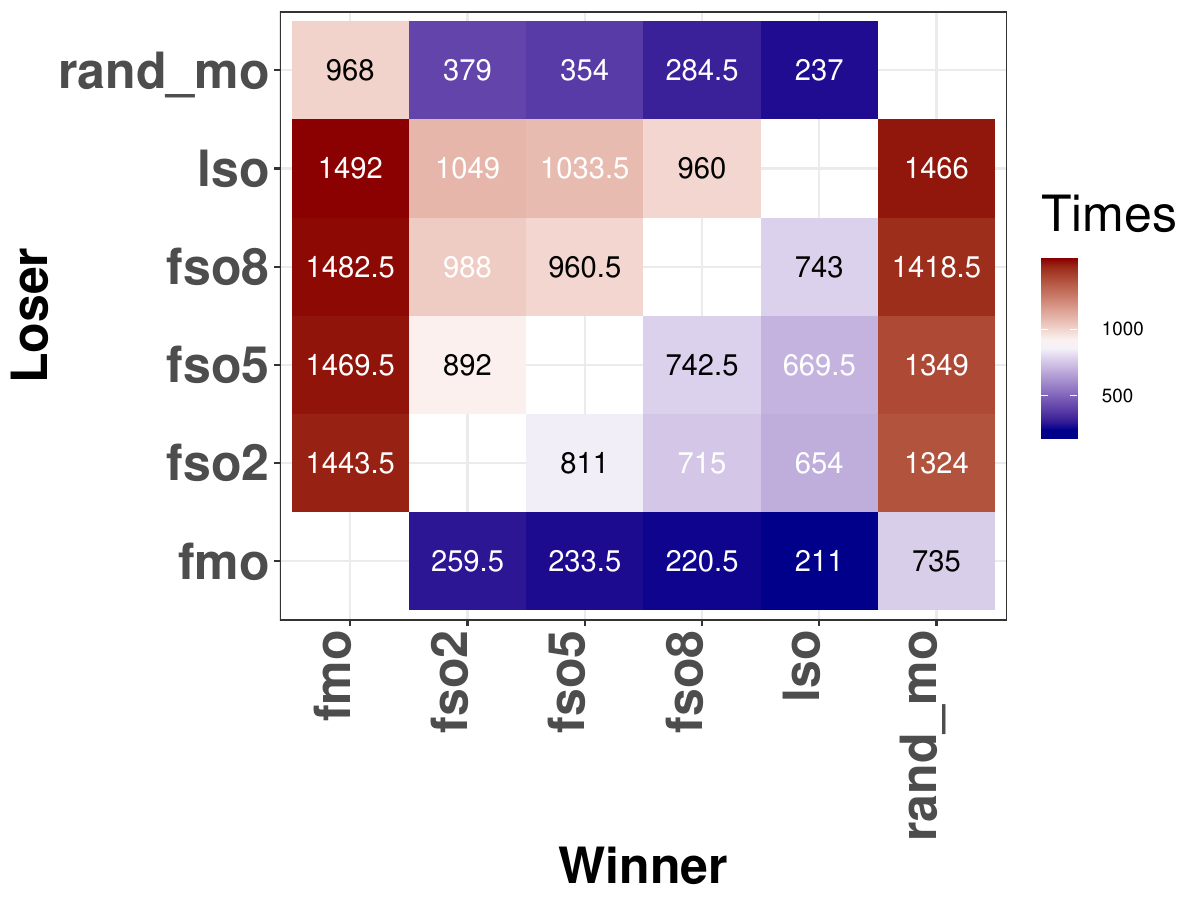}{Aggregated wins and losses on the DRC scenarios obtained over OpenML datasets}{fig:oml-comb-pca-win}{0.45}
	\end{minipage}
\end{figure}

\textbf{Stratified Random Split (SRS): }
To answer the question if a different data splitting technique affects the comparison, we use the stratified technique described in section \ref{sec:semi-sim} which corresponds to the situation that data being more evenly distributed across data sites. 
Figure \ref{fig:oml-comb-stratif-hv} shows the hypervolume plot, from which we can still observe the pattern that the multi-objective candidates perform better in terms of hypervolume, while compared to Figure \ref{fig:hv-oml-comb-pca}, all methods show increased performance under this evenly distributed data scenario across data sites, possibly due to the bonus of evenly distributed data scenario. In Figure \ref{fig:oml-comb-win-stratif}, we compare the wins and losses for each pair of candidates, where in this case, the \fmo\, wins \randmo\, by a larger margin, maybe because the SRS generate a simpler RFMS scenario for the Bayesian Optimization.
\begin{figure}[ht]
	\begin{minipage}[b]{.5\textwidth}
		\mgraph{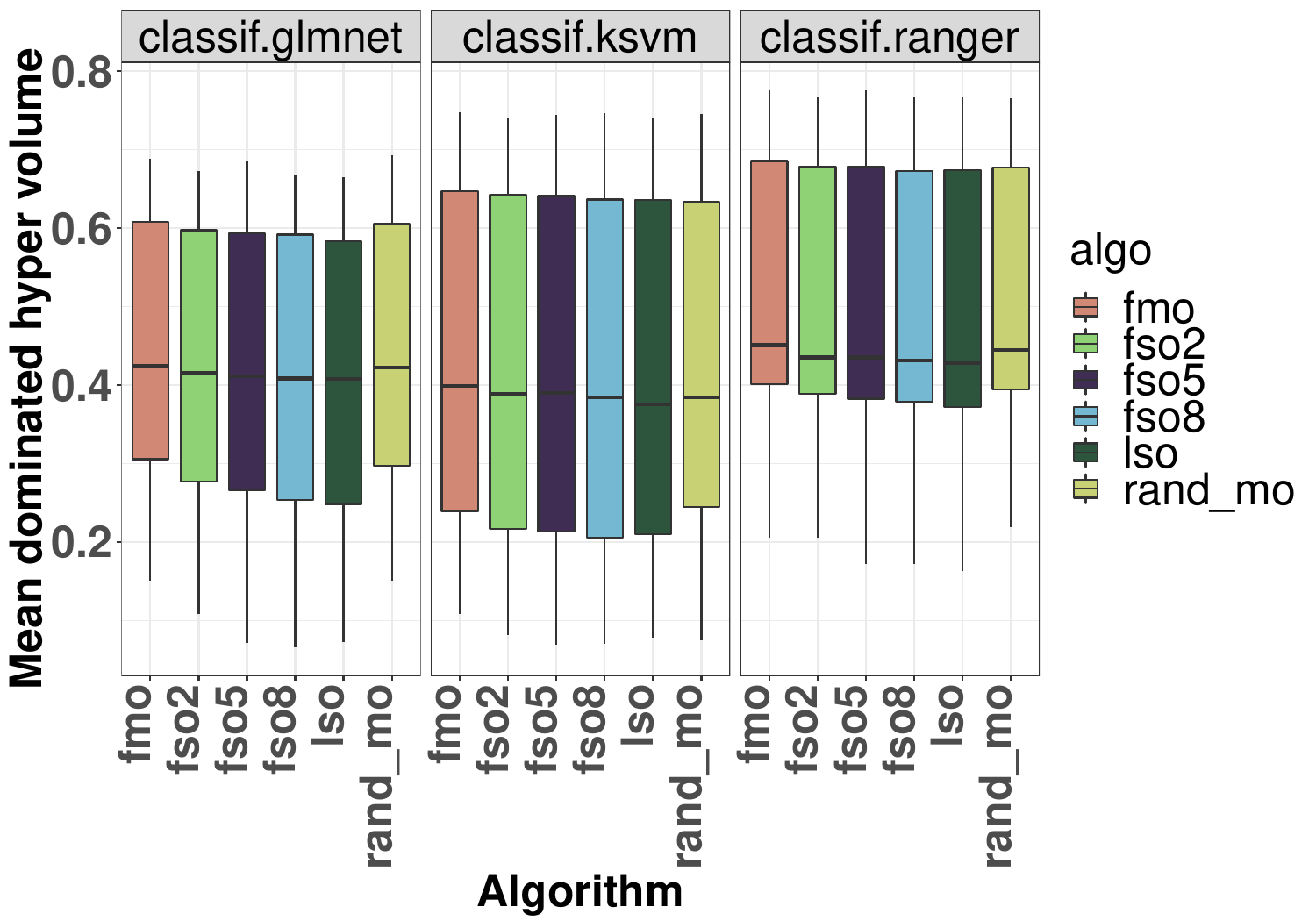}{Aggregated mean dominated hyper-volume under SRS scenario obtained over OpenML datasets}{fig:oml-comb-stratif-hv}{0.3}
	\end{minipage}
	\hfill
	\begin{minipage}[b]{.4\textwidth}
		\mgraph{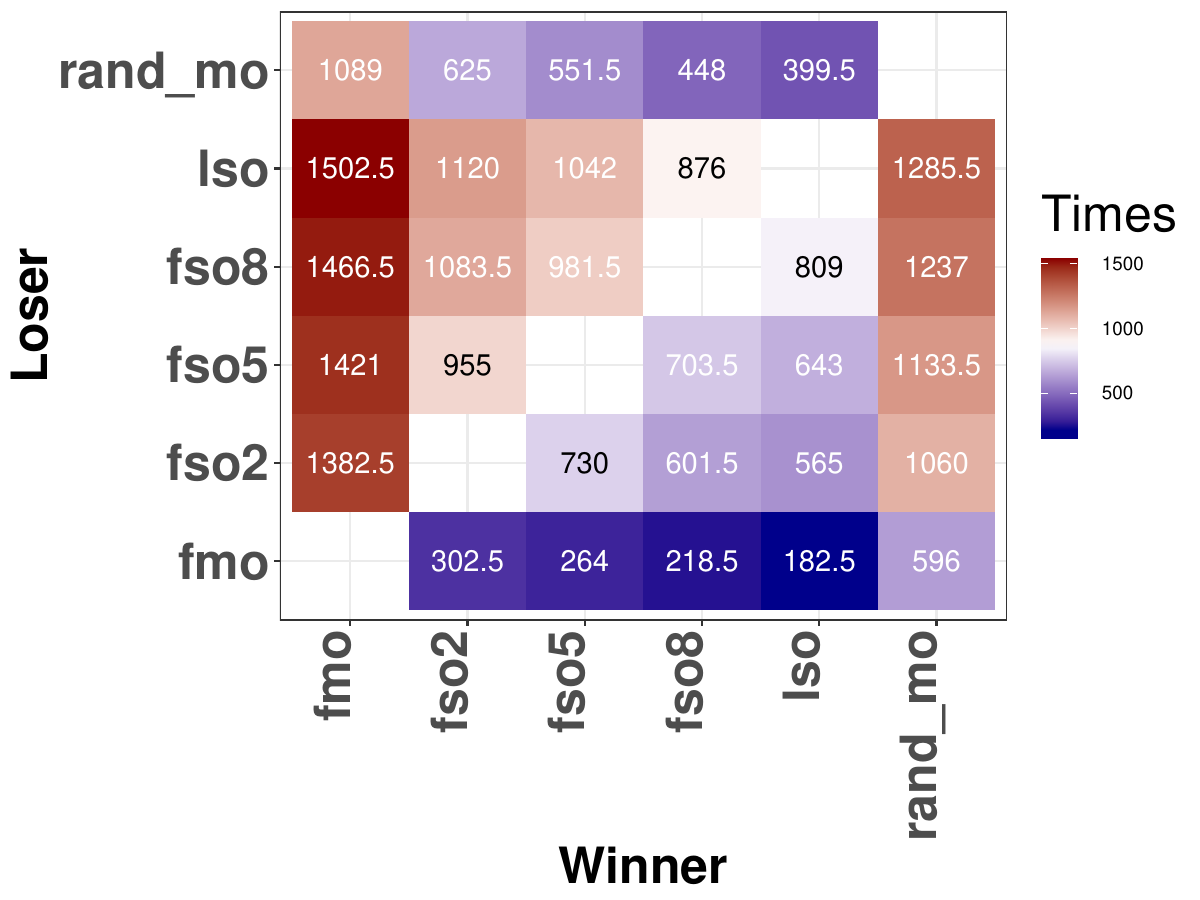}{Aggregated wins and losses over SRS scenario obtained over OpenML datasets}{fig:oml-comb-win-stratif}{0.4}
	\end{minipage}
\end{figure}

 %     \mfig{fig/graph/wins_and_losses_oml3608_stratif.pdf}{win and lossoml3608stratification}{oml-3608-win}{0.4}

  %    \mfig{fig/graph/mean_hypervolumes_oml3608_stratif}{mean mean hyper-volume of oml3608stratification}{oml-3608-stratif-hv}{0.3}

Since close or even identical predicative performance values on a problem can occur for varying machine learning hyperparameters, when the predicative performance is used as the target for Gaussian Process regression, it can create numerical difficulties, so hyperparameter tuning might fail for a particular algorithm, even though we use a nugget value of $1e-6$. Therefore, to get fair comparison, all algorithms are run sequentially over a problem on the same computing node. Only those experiments with all algorithms finished are used for analysis, where in practice, we only get neglectable number of experiments (around 100 out of 1800 experiments, which is 5 percent) within which at least one algorithm is not finished, see Figure \ref{fig:oml-comb-pca-win} and Figure \ref{fig:oml-comb-win-stratif}. The Winner-vs-Loser plots are more effective than carrying out statistical test.
\section{Summary}
% Compared to federated learning, the proposed restricted special case has the following properties:
We introduce a novel learning scenario, Restrictive Federated Model Selection (RFMS), which could play an important role in clinical research, where privacy sensitive immobile high dimensional data is differently distributed among various data sites, in which case federated learning is not applicable due to a lack of access to data from all data sites to be used for training. RFMS is a model selection process in this scenario, with the aim to obtain a model that generalizes comparably well across data sites with potential different distributions. Compared to Federated Learning, RFMS can be carried out in an asynchronous fashion, which is not communication hungry compared to standard federated learning and much easier to be deployed. Additionally, the amount of information that needs to be transferred for each query is comparatively small which takes less efforts to be deployed. 

As an initial investigation, we compare various methods for model selection and hyper-parameter tuning using Bayesian Optimization. Empirical results from various data sources indicate that Federated Multi-objective Bayesian Optimization compares favorably against other single objective candidates as well as multi-objective random search, in terms of better generalization across data sites.
%In order to start the investigation into this problem setting, we compare various methods for model selection and hyper-parameter tuning using Bayesian Optimization, safe guarded by multi-objective random search as a reference to evaluate how effective the Bayesian Optimization works. Our results indicate that multi-objective hyperparameter tuning compares favourably to other single objective baselines and random search.
\section*{Acknowledgement}
This work was supported by Deutsche Forschungsgemeinschaft (DFG), Project RA\,870/7-1 and and BI\,1902/1-1. The authors thank Janek Thomas, Philip Probst and Martin Binder for helpful suggestions.
\bibliographystyle{spmpsci}
\bibliography{references}
\end{document}